\documentclass{article}
\usepackage{spconf,amsmath,graphicx}

\usepackage{xcolor}

\usepackage{amssymb}
\usepackage{pifont}
\usepackage{hyperref}
\usepackage{amsmath}
\usepackage{multirow}

\usepackage{enumitem}
\setlist{nosep, leftmargin=14pt}

\usepackage{mwe} 

\title{Fitting Skeletal Models via Graph-based Learning}

\name{Nicolás Gaggion$^{\star}$ \qquad Enzo Ferrante$^{\star}$ \qquad Beatriz Paniagua$^{\dagger}$ \qquad Jared Vicory$^{\dagger}$}

\address{$^{\star}$ CONICET \\
    $^{\dagger}$ Kitware, Inc. }

\begin{document}
%
\maketitle
\begin{abstract}
Skeletonization is a popular shape analysis technique that models an object's interior as opposed to just its boundary. Fitting template-based skeletal models is a time-consuming process requiring much manual parameter tuning. Recently, machine learning-based methods have shown promise for generating s-reps from object boundaries. In this work, we propose a new skeletonization method which leverages graph convolutional networks to produce skeletal representations (s-reps) from dense segmentation masks. The method is evaluated on both synthetic data and real hippocampus segmentations, achieving promising results and fast inference.
\end{abstract}


\section{Introduction}
\label{sec:intro}
\begin{figure}[!t]
	\centering
	\includegraphics[width=0.9\columnwidth]{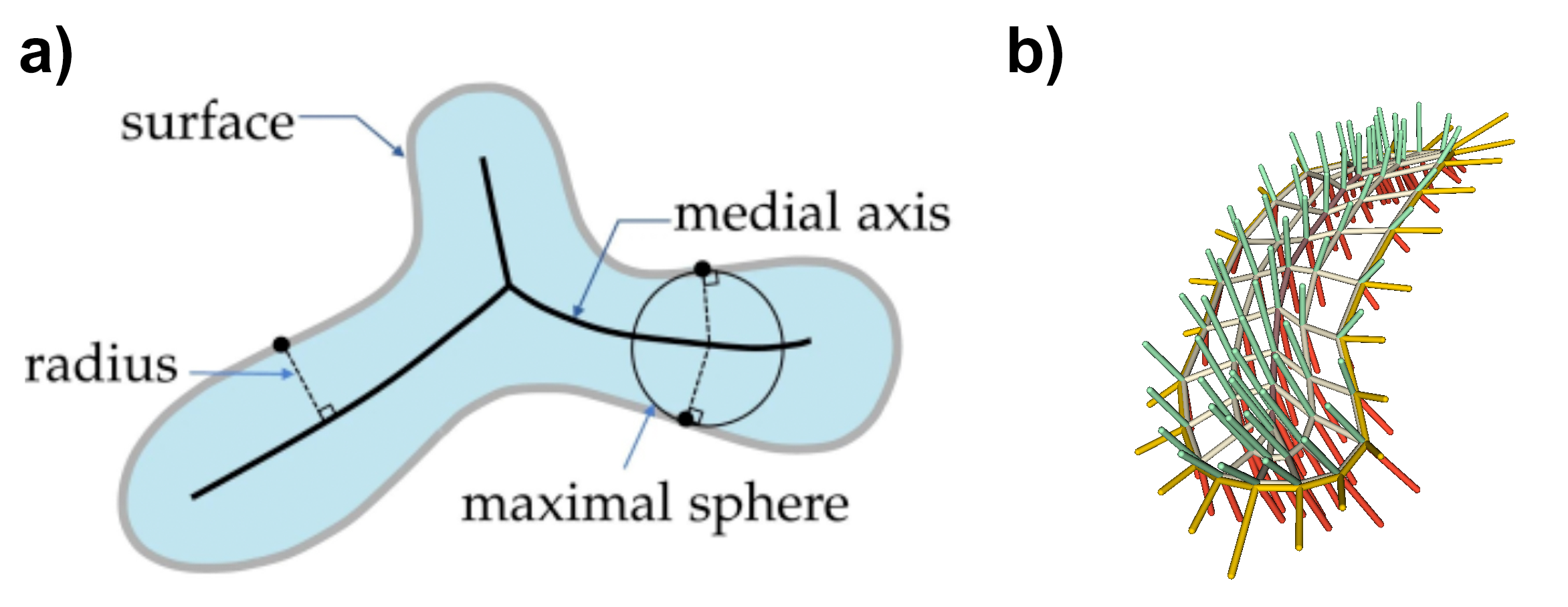}
	\caption{(Left) Medial axis for a 2D shape, (Right) s-rep for a hippocampus
	         surface with yellow lines as the spoke vectors.}
	\label{fig:teaser}
\end{figure}

Skeletonization has been a powerful approach for modeling anatomical structures because they model both the object's boundary and its interior, as compared with simpler models such as calculating densely sampled boundary landmarks. Historically, a popular way to define the skeleton of an object is through Blum's medial axis transform~\cite{blum1967transformation-mat-org} (MAT)(fig. \ref{fig:teaser}.b). The MAT consists of a set of points and associated radii, which we call spokes, that form the set of maximally inscribed spheres inside the shape. MAT-based models have been used for a wide range of applications~\cite{tagliasacchi20163d-skel-survey} such as segmentation, registration and statistics of object shape.

The main limitation of medial models is that they have a tendency to amplify small-scale noise on an object's boundary, resulting in inconsistencies in skeleton location and topology across a population. This makes this representation hard to apply to real life problems where objects from the same population are usually highly variable. This limitation has led to multiple variations of the MAT~\cite{yan2018voxel-matapprox, skel3d-wu2015-DPC}. In particular, \textit{skeletal representations (s-reps)}~\cite{pizer2020-srep-object}(fig. \ref{fig:teaser}, right) are a class of discrete skeletal representations that relate to the MAT but have a fixed topology and can achieve consistent sampling across a population. This is done by fitting a template s-rep to an object via optimization~\cite{liu2021-srep-fitting} rather than direct computation from the object's boundary. Having a fixed template that is optimized to fit each individual object yields improved consistency, correspondence and resistance to local noise. The optimization process has constraints that allow the final object be nearly medial, including enforcing points on the skeleton to be approximately equidistant from the top and bottom of the object's surface, and the radii associated to these points to be nearly orthogonal to the boundary. The optimization can be slow and often requires manual template generation and parameter tuning when applied to a new data set.

Machine learning methods for skeletonizing images and shapes are a relatively recent line of research which have shown promise in robustly computing s-reps. Earlier learning-based methods were primarily focused on extracting 2D skeletons from images~\cite{skel2d-shen2017deepskeleton,skel2d-nguyen2021u}. In contrast, there is less work on learning-based skeletonization of 3D objects, partially due to their increased complexity and to the lack of a benchmark dataset for training. This led to the development of point-based methods like Point2Skeleton~\cite{skel3d-p2skel-lin2021} which utilizes PointNet++~\cite{qi2017pointnet++} as a point encoder and tries to predict weights on input points to generate the skeleton as a convex combination of the inputs in a manner similar to~\cite{skel3d-chen2020unsupervised}. Our previous work~\cite{10230505} adapted a point-based approach with additional medialness constraints to produce medial skeletons from 3D surfaces.

Recent advances in graph-based neural networks have made their direct application to deriving s-reps from images possible. HybridGNet \cite{gaggion2022improving} combined a convolutional encoder with a graph-based decoder to segment an image by estimating a contour with a fixed number of points. HybridVNet \cite{gaggion2023multiview} has recently extended this approach to 3D, allowing the direct extraction of volumetric meshes of a fixed topology from a 3D image. In this work, we build on top of this approach to directly predict a skeleton from dense binary segmentation masks, by conceiving s-reps as graph structures.

\section{Methodology}
\label{sec:method}

\subsection{Skeletal Representations}

An s-rep consists of a grid of points on the skeleton and a set of vectors emanating from the skeleton to the boundary called spokes (see \ref{fig:teaser}.b.), that explicitly represent both the object's full interior and surface. Because an s-rep has a fixed grid structure of corresponding points on both the skeleton and boundary, we can easily use it to derive the volumetric graph representation needed by the method described in section ~\ref{subsec:hybridvnet}. By connecting each quad of neighboring points on the skeletal surface and the corresponding quad on the object boundary, we form a single volumetric element which is then decomposed into tetrahedra. While in this work we only use one element to connect the skeleton to the boundary, this could be made more dense by subdividing based on distance along the spokes. Others have used a similar approach to generate models for finite element analysis from s-reps~\cite{crouch2007automated}. 

\subsection{S-reps via HybridVNet}
\label{subsec:hybridvnet}

\begin{figure*}
    \centering
    \includegraphics[width=\linewidth]{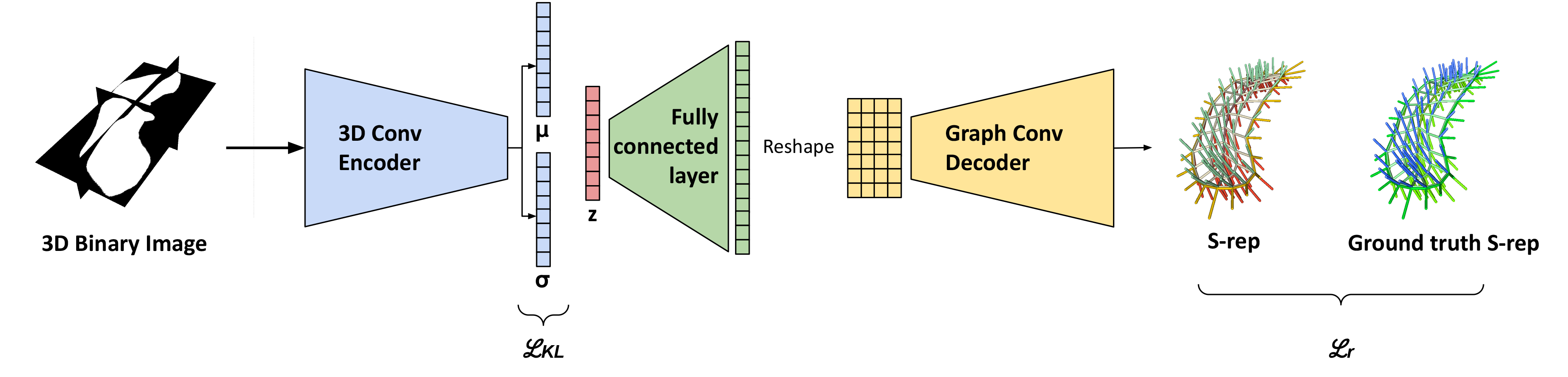}
    \caption{\textbf{Model architecture:} The presented model utilizes a variational encoder-decoder architecture to create a graph representation of an s-rep derived from a binary input image. The encoder comprises a 3D convolutional neural network, producing $\boldsymbol{\mu}$ and $\boldsymbol{\sigma}$ vectors which are  sampled, yielding a latent representation denoted as $\mathbf{z}$. This latent code is subsequently goes through a fully connected layer and is reshaped to establish the primary node attributes for the graph convolutional decoder. Leveraging these initial node attributes, the decoder generates the conclusive graph representation of the s-rep.}
    \label{fig:model_arch}
\end{figure*}

HybridVNet \cite{gaggion2023multiview} employs a hybrid encoder-decoder architecture tailored for generating meshes directly from images. Figure \ref{fig:model_arch} shows the proposed HybridVNet's single view architecture, that encompasses a 3D convolutional encoder to encode input images and derive a latent space representation of the target object. The resulting encoded representation is further processed through a fully-connected layer and reshaped to initialize features for the subsequent decoder stage.

In contrast to a typical convolutional decoder, HybridVNet adopts spectral graph convolutional layers \cite{defferrard2016convolutional} for transforming the latent representation into the desired graph structure representative of s-reps. The decoder comprises five graph convolutional layers interleaved with rectified linear unit (ReLU) nonlinearities and Layer Normalization. Notably, the removal of graph unpooling layers distinguishes this modified architecture, considering the smaller output resolution and the intricate relationships between edges necessitated by the complex graph structure in s-reps.

More formally, the s-rep HybridVNet is implemented as a variational autoencoder \cite{kingma2013auto} where the convolutional encoder $E: \mathcal{S} \rightarrow \mathbb{R}^{2,d}$ takes an input binary segmentation mask $s\in \mathcal{S}$ and outputs the parameters (mean and variance) of a multivariate Gaussian distribution as $\mu, \sigma=E(s)$. A latent code $z$ is then sampled from the distribution $z \sim \mathcal{N}(\mu, \sigma)$ following the reparametrization trick. $z$ is then reshaped and enters a graph convolutional decoder $D:\mathbb{R}^d \rightarrow \mathcal{G}$, producing a graph $G=D(z) \in \mathcal{G}$ modelling the s-rep. The graph s-rep is defined as $G=<V,A,X>$ where $V$ is the set of nodes representing the skeleton and the surface vertices, $A$ is the adjacency matrix of the template s-rep where $A_{i,j}=1$ when there is an edge between nodes $(i,j)$ and 0 otherwise. 
Finally, $X\in \mathbb{R}^{3,|V|}$ is a function assigning a 3D coordinate to every node $v \in |V|$.

The loss function for the network is a weighted sum comprised of the following terms:

\begin{itemize}
  \item $\mathcal{L}_r$ (Reconstruction Loss): Computed as the mean squared error (MSE) of vertex positions, facilitating the fidelity of generated s-reps.
  \item $\mathcal{L}_{KL}$ (KL Divergence Loss): Imposes a unit Gaussian prior on the latent distribution, guiding the network's learning towards a structured latent space.
\end{itemize}

\section{Results}
\label{sec:experiments}

\subsection{Experiment Setup}

\subsubsection{Data Collection and Partitioning}

We used a data set of synthetic s-reps and another of s-reps of hippocampi in order to benchmark the proposed algorithm.

\textbf{Synthetic Dataset:} This dataset comprised 5000 randomly simulated binary ellipsoid images with analytically derived s-reps. Starting from a base ellipsoid and s-rep which are axis aligned, we first applied random scale factors sampled from the normal distribution $\mathcal{N}(1,0.15)$ to each axis independently. We then deform the ellipsoid by bending the long axis by angles sampled from $\mathcal{N}(\frac{\pi}{3},\frac{\pi}{8})$ and twisting by angles sampled from $\mathcal{N}(\frac{\pi}{6},\frac{\pi}{8})$.  This dataset was split randomly, allocating 80\% for training/validation and 20\% for testing.

\textbf{Hippocampus Dataset:} This dataset consisted of 175 pairs of binary images segmented from magnetic resonance imaging (MRI) with associated s-reps obtained using the ellipsoid template warping method described in \cite{liu2021-srep-fitting}. This dataset was divided equally into training/validation and test partitions. The training set was further subdivided into five splits of 10\%, 20\%, 30\%, 40\%, and 50\% to explore the impact of varying training data size. For training the models, we defined an epoch to have the same number of iterations (900) as the synthetic experiment.

\subsubsection{Model Training}

The HybridVNet architecture was trained from scratch for a fixed number of 50 epochs, retaining the best model based on an internal validation split for subsequent testing. For fine-tuning, the best model from the synthetic dataset and a maximum of number of 10 training epochs was set. Online data augmentation techniques, including random rotations and scaling in the three spatial dimensions, were applied to the data.

Hyperparameters, determined via a grid search, included weighting factors ($\lambda_{r} = 1$, $\lambda_{KL} = 1e-3$) for the reconstruction loss ($\mathcal{L}_{r}$) and KL divergence loss ($\mathcal{L}_{KL}$), a learning rate of $1e-4$ with decay set at $0.99$ per epoch, and a batch size of $4$, accounting for GPU memory constraints.

\subsubsection{Evaluation Metrics}

Model performance was assessed using multiple metrics on both the raw point positions and s-rep-related metrics:

\begin{itemize}
    \item \textbf{Positional Metrics:} Mean Average Error (MAE), Mean Squared Error (MSE), and Root Mean Squared Error (RMSE) calculated based on positional coordinates in physical space for the skeleton and boundary points.
    \item \textbf{Skeleton-Based Metrics:}
    \begin{itemize}
        \item \textit{Medialness:} Average ratio between the lengths of top and bottom spokes for each skeletal point.
        \item \textit{Angles:} Average angle between corresponding spokes between two s-reps.
        \item \textit{Orthogonality:} Average angle between spoke directions and boundary normals, requiring the utilization of the encompassing surface mesh of the structure.
    \end{itemize}
\end{itemize}

The angles metric necessitated a direct comparison with the ground-truth s-rep, while orthogonality calculations relied on the surface mesh information.

\subsection{Synthetic experiment}    

\begin{table}[h]
\caption{Synthetic dataset results. Mean (Std)}
\label{simulated_table}
\resizebox{\linewidth}{!}{
\begin{tabular}{cccccc}
\hline
\textbf{MSE $\downarrow$} & \textbf{MAE $\downarrow$} & \textbf{RMSE $\downarrow$} & \textbf{Medialness $\uparrow$} & \textbf{Angle} & \textbf{Orthogonality} \\ \hline
0.07 (0.05)  & 0.21 (0.06)  & 0.26 (0.07)   & 0.99 (0.03)         & 0.17 (0.05)    & 0.32 (0.10)            \\ \hline
\end{tabular}}
\end{table}

Table \ref{simulated_table} shows results on the synthetic ellipsoid data. The point-based metrics show strong performance in producing predicted models close to the analytically derived s-reps on both the skeleton and boundary. The angle measure shows good agreement between the spoke directions of the GT and predicted models. This model serves as the base for the finetuning experiments on clinical Hippocampus data.

\subsection{Hippocampus experiment}

\begin{table}[h]
\caption{Hippocampus dataset results. Mean (Std)}
\label{hippocampus_table}
\resizebox{\linewidth}{!}{
\begin{tabular}{cc|cccccc}
\hline
Model                      & Percentage & MSE $\downarrow$        & MAE $\downarrow$        & RMSE $\downarrow$      & Medialness $\uparrow$  & Angle       & \multicolumn{1}{l}{Orthogonality} \\ \hline
GT Reference               & -          & -           & -           & -           & 0.99 (0.03) & -           & 0.47 (0.04                        \\ \hline
\multirow{5}{*}{Finetuned} & 10         & 0.47 (0.29) & 0.50 (0.14) & 0.66 (0.19) & 0.97 (0.01) & 0.21 (0.05) & 0.48 (0.04)                       \\
                           & 20         & 0.43 (0.28) & 0.48 (0.14) & 0.63 (0.19) & 0.98 (0.01) & 0.2 (0.05)  & 0.48 (0.04)                       \\
                           & 30         & 0.45 (0.29) & 0.49 (0.14) & 0.65 (0.19) & 0.98 (0.01) & 0.2 (0.04)  & 0.47 (0.04)                       \\
                           & 40         & 0.44 (0.25) & 0.49 (0.12) & 0.64 (0.17) & 0.98 (0.01) & 0.19 (0.04) & 0.47 (0.04)                       \\
                           & 50         & 0.42 (0.33) & 0.46 (0.14) & 0.62 (0.21) & 0.98 (0.01) & 0.2 (0.04)  & 0.47 (0.03)                       \\ \hline
From Scratch               & 50         & 0.46 (0.34) & 0.48 (0.15) & 0.64 (0.22) & 0.99 (0.02) & 0.17 (0.05) & 0.48 (0.04)                       \\ \hline
\end{tabular}}
\end{table}

Table \ref{hippocampus_table} show results from the hippocampus experiment in which the model trained on the synthetic data was finetuned with progressively more hippocampus examples. The results indicate relatively favorable performance in estimating skeletal features. Particularly in capturing angle differences between the predicted and ground truth models, consistent preservation of medialness across experiments and comparable orthogonality metrics to existing fitting methods (used as ground truth) suggest the proposed approach's validity and feasibility in generating s-reps from real hippocampal data.

\section{Conclusion}
\label{sec:conclusion}

In this work we adapt a hybrid convolutional/graph neural network, initially proposed for graph-based anatomical segmentation, to generate skeletal representations from object boundaries represented as binary images. As a benefit over previous learning-based skeletonization approaches, this method directly encodes the connections between the object skeleton and boundary to encourage the result to have a well-behaved skeletal structure in terms of medialness and boundary orthogonality. The results on a dataset of clinical objects shows similar performance in this metrics to previous s-rep fitting approaches based on deformable template optimization in significantly less time. The inference time for the forward pass of the network is 0.24 seconds per input image on an NVIDIA RTX A5000 GPU or 2.5 seconds per image on an Intel(R) Core(TM) i7-7700 CPU operating at 3.60GHz, while optimization-based approaches take at least several minutes and some times significantly longer.

There are several avenues for further improvement of the preliminary work presented here. While the method seems to produce s-reps with reasonably good structures, we are not currently directly encoding desirable traits such as medialness or spoke/boundary orthogonality into the loss functions used to train the models. This could further improve the results, particularly in cases where training data is limited.

\section{Compliance with Ethical Standards}
\label{sec:compliance-ethical}
The hippocampus data was provided by Martin Styner, UNC Neuro Image Analysis Laboratory (NIRAL). The study was performed according to a protocol approved by the institutional review board at the relevant institutions.


\bibliographystyle{IEEEbib}
\bibliography{refs}

\begin{thebibliography}{10}

\bibitem{blum1967transformation-mat-org}
Harry Blum,
\newblock ``A transformation for extracting new descriptions of shape,''
\newblock {\em Models for the perception of speech and visual form}, pp. 362--380, 1967.

\bibitem{tagliasacchi20163d-skel-survey}
Andrea Tagliasacchi, Thomas Delame, Michela Spagnuolo, Nina Amenta, and Alexandru Telea,
\newblock ``3d skeletons: A state-of-the-art report,''
\newblock in {\em Computer Graphics Forum}. Wiley Online Library, 2016, vol.~35, pp. 573--597.

\bibitem{yan2018voxel-matapprox}
Yajie Yan, David Letscher, and Tao Ju,
\newblock ``Voxel cores: Efficient, robust, and provably good approximation of 3d medial axes,''
\newblock {\em ACM Transactions on Graphics (TOG)}, vol. 37, no. 4, pp. 1--13, 2018.

\bibitem{skel3d-wu2015-DPC}
Shihao Wu, Hui Huang, Minglun Gong, Matthias Zwicker, and Daniel Cohen-Or,
\newblock ``Deep points consolidation,''
\newblock {\em ACM Transactions on Graphics (ToG)}, vol. 34, no. 6, pp. 1--13, 2015.

\bibitem{pizer2020-srep-object}
Stephen~M Pizer, Junpyo Hong, Jared Vicory, Zhiyuan Liu, JS~Marron, Hyo-young Choi, James Damon, Sungkyu Jung, Beatriz Paniagua, J{\"o}rn Schulz, et~al.,
\newblock ``Object shape representation via skeletal models (s-reps) and statistical analysis,''
\newblock in {\em Riemannian Geometric Statistics in Medical Image Analysis}, pp. 233--271. Elsevier, 2020.

\bibitem{liu2021-srep-fitting}
Zhiyuan Liu, Junpyo Hong, Jared Vicory, James~N Damon, and Stephen~M Pizer,
\newblock ``Fitting unbranching skeletal structures to objects,''
\newblock {\em Medical Image Analysis}, vol. 70, pp. 102020, 2021.

\bibitem{skel2d-shen2017deepskeleton}
Wei Shen, Kai Zhao, Yuan Jiang, Yan Wang, Xiang Bai, and Alan Yuille,
\newblock ``Deepskeleton: Learning multi-task scale-associated deep side outputs for object skeleton extraction in natural images,''
\newblock {\em IEEE Transactions on Image Processing}, vol. 26, no. 11, pp. 5298--5311, 2017.

\bibitem{skel2d-nguyen2021u}
Nam~Hoang Nguyen,
\newblock ``U-net based skeletonization and bag of tricks,''
\newblock in {\em Proceedings of the IEEE/CVF International Conference on Computer Vision}, 2021, pp. 2105--2109.

\bibitem{skel3d-p2skel-lin2021}
Cheng Lin, Changjian Li, Yuan Liu, Nenglun Chen, Yi-King Choi, and Wenping Wang,
\newblock ``Point2skeleton: Learning skeletal representations from point clouds,''
\newblock in {\em Proceedings of the IEEE/CVF Conference on Computer Vision and Pattern Recognition}, 2021, pp. 4277--4286.

\bibitem{qi2017pointnet++}
Charles~Ruizhongtai Qi, Li~Yi, Hao Su, and Leonidas~J Guibas,
\newblock ``Pointnet++: Deep hierarchical feature learning on point sets in a metric space,''
\newblock {\em Advances in neural information processing systems}, vol. 30, 2017.

\bibitem{skel3d-chen2020unsupervised}
Nenglun Chen, Lingjie Liu, Zhiming Cui, Runnan Chen, Duygu Ceylan, Changhe Tu, and Wenping Wang,
\newblock ``Unsupervised learning of intrinsic structural representation points,''
\newblock in {\em Proceedings of the IEEE/CVF Conference on Computer Vision and Pattern Recognition}, 2020, pp. 9121--9130.

\bibitem{10230505}
Ninad Khargonkar, Beatriz Paniagua, and Jared Vicory,
\newblock ``Skeletal point representations with geometric deep learning,''
\newblock in {\em 2023 IEEE 20th International Symposium on Biomedical Imaging (ISBI)}, 2023, pp. 1--5.

\bibitem{gaggion2022improving}
Nicol{\'a}s Gaggion, Lucas Mansilla, Candelaria Mosquera, Diego~H Milone, and Enzo Ferrante,
\newblock ``Improving anatomical plausibility in medical image segmentation via hybrid graph neural networks: applications to chest x-ray analysis,''
\newblock {\em IEEE Transactions on Medical Imaging}, vol. 42, no. 2, pp. 546--556, 2022.

\bibitem{gaggion2023multiview}
Nicol\'as Gaggion, Benjamin~A. Matheson, Yan Xia, Rodrigo Bonazzola, Nishant Ravikumar, Zeike~A. Taylor, Diego~H. Milone, Alejandro~F. Frangi, and Enzo Ferrante,
\newblock ``Multi-view hybrid graph convolutional network for volume-to-mesh reconstruction in cardiovascular mri,''
\newblock {\em Arxiv}, 2023.

\bibitem{crouch2007automated}
Jessica~R Crouch, Stephen~M Pizer, Edward~L Chaney, Yu-Chi Hu, Gig~S Mageras, and Marco Zaider,
\newblock ``Automated finite-element analysis for deformable registration of prostate images,''
\newblock {\em IEEE transactions on medical imaging}, vol. 26, no. 10, pp. 1379--1390, 2007.

\bibitem{defferrard2016convolutional}
Micha{\"e}l Defferrard, Xavier Bresson, and Pierre Vandergheynst,
\newblock ``Convolutional neural networks on graphs with fast localized spectral filtering,''
\newblock {\em arXiv preprint arXiv:1606.09375}, 2016.

\bibitem{kingma2013auto}
Diederik~P Kingma and Max Welling,
\newblock ``Auto-encoding variational bayes,''
\newblock {\em arXiv preprint arXiv:1312.6114}, 2013.

\end{thebibliography}

\end{document}